\documentclass[letterpaper, 10 pt, conference]{ieeeconf}  

\IEEEoverridecommandlockouts                              

\usepackage{graphicx}
\usepackage{amssymb}
\usepackage{booktabs}

\usepackage[pagebackref,breaklinks,colorlinks]{hyperref}

\usepackage{graphicx}
\usepackage{tikz}
\usepackage{comment}
\usepackage{amsmath,amssymb,bm} 
\usepackage{tabularx}
\usepackage{multirow}

\usepackage{booktabs}
\usepackage{xcolor}

\usepackage{threeparttable}
\usepackage{adjustbox}
\usepackage{eqparbox}
\usepackage{arydshln}
\usepackage{wrapfig}
\usepackage{interval}
\usepackage{algorithm}
\usepackage{algpseudocode}

\makeatletter
\newcommand{\multiline}[1]{%
  \begin{tabularx}{\dimexpr\linewidth-\ALG@thistlm}[t]{@{}X@{}}
    #1
  \end{tabularx}
}
\makeatother

\newcommand{\distas}[1]{\mathbin{\overset{#1}{\kern\z@\sim}}}%
\newsavebox{\mybox}\newsavebox{\mysim}
\newcommand{\distras}[1]{%
	\savebox{\mybox}{\hbox{\kern3pt$\scriptstyle#1$\kern3pt}}%
	\savebox{\mysim}{\hbox{$\sim$}}%
	\mathbin{\overset{#1}{\kern\z@\resizebox{\wd\mybox}{\ht\mysim}{$\sim$}}}%
}

\title{Monocular Visual 8D Pose Estimation for Articulated Bicycles and Cyclists}

\author{Eduardo R. Corral-Soto\textsuperscript{*}, Yang Liu, Yuan Ren, Bai Dongfeng, Liu Bingbing 
	\thanks{All authors are with Huawei Noah's Ark Lab, Canada at the time of writing.  
		{\tt\small \{eduardo.corral.soto, yang.liu9, yuan.ren3, baidongfeng, liu.bingbing  \}@huawei.com}}%
        \thanks{*Main author and R\&D contributor}%
}

\begin{document}

\maketitle
\thispagestyle{empty}
\pagestyle{empty}

\begin{abstract}
In Autonomous Driving, cyclists belong to the safety-critical class of Vulnerable Road Users (VRU), and accurate estimation of their pose is critical for cyclist crossing intention classification, behavior prediction, and collision avoidance. Unlike rigid objects, articulated bicycles are composed of movable rigid parts linked by joints and constrained by a kinematic structure. 6D pose methods can estimate the 3D rotation and translation of rigid bicycles, but 6D becomes insufficient when the steering/pedals angles of the bicycle vary. That is because: 1) varying the articulated pose of the bicycle causes its 3D bounding box to vary as well, and 2) the 3D box orientation is not necessarily aligned to the orientation of the steering which determines the actual intended travel direction. In this work, we introduce a method for category-level 8D pose estimation for articulated bicycles and cyclists from a single RGB image. Besides being able to estimate the 3D translation and rotation of a bicycle from a single image, our method also estimates the rotations of its steering handles and pedals with respect to the bicycle body frame. These two new parameters enable the estimation of a more fine-grained bicycle pose state and travel direction. Our proposed model jointly estimates the 8D pose and the 3D Keypoints of articulated bicycles, and trains with a mix of synthetic and real image data to generalize on real images. We include an evaluation section where we evaluate the accuracy of our estimated 8D pose parameters, and our method shows promising results by achieving competitive scores when compared against state-of-the-art category-level 6D pose estimators that use rigid canonical object templates for matching.
\end{abstract}

\section{Introduction}
\label{sec:intro}


\begin{figure} 
	\centering	
		\includegraphics[width=1.0\columnwidth, trim={1cm 16.5cm 1cm 0cm},clip]{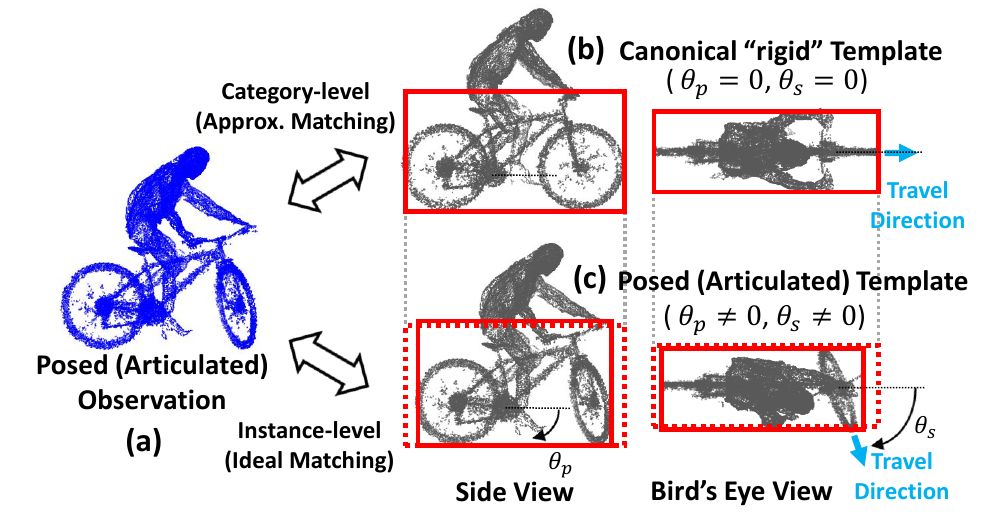} 		
	\caption{ Category-level 6D pose estimators can find an approximate match between an observed \emph{articulated posed} cyclist (a) and a \emph{rigid canonical} cyclist (b) template. But that introduces two problems: \textbf{1)} A mis-match between the 3D bounding boxes of the canonical rigid template (c)-dashed and the articulated posed cyclist (c)-solid. \textbf{2)} 
The cyclist travel directions (cyan arrows) are quite different from each other (c) v/s (b), although the dashed and solid boxes appear to be aligned. Ideally, an observed articulated posed cyclist (a) should be matched with an \emph{articulated posed} template (c), as done in instance-level 6D methods.   
	\label{Cano_vs_Posed_Mismatch} }
\end{figure}

\textbf{Context:} 
In Autonomous Driving, cyclists are considered Vulnerable Road Users (VRU), and accurate estimation of their pose is critical for cyclist crossing intention classification ~\cite{abadi2023detection}, ~\cite{kress2019pose}, for predicting their behavior ~\cite{gu2014recognition}, ~\cite{cho2010vision}, to avoid collisions ~\cite{gu2014recognition}.\\ 
\textbf{Problem and Motivation:} Existing 6D and 9D pose estimation methods focus on estimating the 3D rotation and translation of \emph{rigid} objects and typically \emph{require either CAD models, 3D pointclouds, or depth maps} as inputs beside RGB images for training and inference. Unlike rigid objects, articulated bicycles are composed of movable rigid parts (i.e. body frame, steering, pedals) linked by joints, and constrained by a kinematic structure. Recent category-level 6D pose estimation methods ~\cite{lin2024sam, wen2024foundationpose, labbe2022megapose} estimate an initial coarse 6D pose of a rigid object followed by a refinement. They obtain an initial 6D pose by finding the best matching/alignment between an observed object 3D pointcloud and 3D pointclouds from a set of pre-generated templates also from $\emph{rigid}$ objects of the same category. These category-level pose estimation methods can still work with articulated bicycles and cyclists by finding the best approximate match between the observed articulated \emph{posed} cyclist (see Fig. \ref{Cano_vs_Posed_Mismatch}(a)) and a \emph{rigid} \emph{canonical} cyclist (Fig. \ref{Cano_vs_Posed_Mismatch}(b)). An ideal solution would be to match each observed articulated posed cyclist with an articulated \emph{posed} template (Fig. \ref{Cano_vs_Posed_Mismatch}(c)) as done in instance-level pose estimation methods, but \textbf{pre-generating a large, diverse set of articulated posed hypotheses} (each one with its own 6D pose, plus the two additional Degrees of Freedom (DoF) for the steering and pedals angles) would become a \textbf{bottle neck} in terms of memory and run-time as the number of hypotheses increases. On the other hand, \textbf{varying the articulated pose} of the bicycle/cyclist causes the definition of its \textbf{3D bounding box to vary as well}. This causes a mis-match between the 3D bounding boxes from the canonical \emph{rigid} templates (red dashed box in Fig. \ref{Cano_vs_Posed_Mismatch}(c)) and the 3D bounding boxes from articulated \emph{posed} cyclists (red solid box in Fig. \ref{Cano_vs_Posed_Mismatch}(c)). Moreover, \textbf{the actual cyclist travel directions} (cyan arrows) between Fig. \ref{Cano_vs_Posed_Mismatch} (b) and (c) are \textbf{quite different} from each other, although the dashed and solid 3D bounding boxes appear to be reasonably aligned on Fig. \ref{Cano_vs_Posed_Mismatch}. These problems motivated us to obtain more fine-grained pose estimates for articulated bicycles and cyclists. \\
\textbf{Our contributions} can be summarized as follows: 
\begin{itemize}
  \item We introduce the task of category-level 8D pose estimation for \emph{articulated} bicycles from a single RGB image. Besides being able to estimate the 3D translation and rotation of a bicycle, our method also estimates: 1) the rotation of its steering handles and pedals with respect to the bicycle body frame, and 2) the bicycle 3D Keypoints.
  \item For this task, we propose an end-to-end regression model design that splits the 8D pose estimation problem into: 1) the regression of the 3D bicycle part rotations using features learned via a Keypoint branch that performs self-attention on a bicycle sub-image that includes structural and appearance information from key bicycle parts,  and 2) a Translation branch that uses these shared features to regress the 3D bicycle translation.
  \item Our model regresses bicycle canonical 3D Keypoints that are reposed by a parametric 3D bicycle model using the regressed 8D pose to produce a set of 3D Keypoints from the observed input image. Their 2D projections are used to help supervise the model during training.
\end{itemize}



\begin{figure*} 
	\centering	
		\includegraphics[width=1.5\columnwidth, trim={0cm 14.0cm 0cm 0cm},clip]{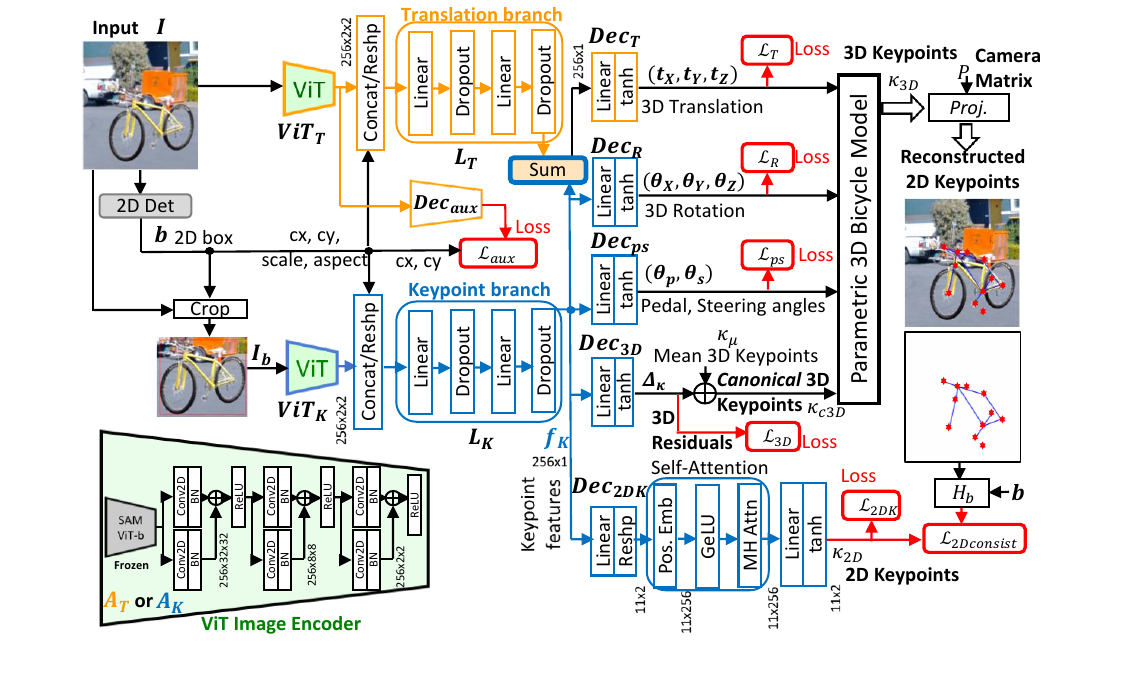}	
	 \caption{figure}{Proposed Monocular Visual Bicycle 8D Pose Estimation Model. The inputs to our model are an RGB image $I$ of a bicycle and its 2D bounding box $b$ obtained from an off-the-shelf 2D object detector. The outputs are the 8D bicycle pose parameters $\{ \theta_p, \theta_s, \theta_{X}, \theta_{Y}, \theta_{Z},  t_X$, $t_Y$, $t_Z \}$, and a set $\kappa_{3D}$ of $11$ regressed bicycle 3D Keypoints (red markers on the right side of the diagram).  
         \label{model_diagram}}
\end{figure*}

\subsection{Related Work}
\label{sec:related_work}

\textbf{6D and 9D Pose Estimation Methods:}
6D pose estimation methods estimate the 3D translation and rotation of \emph{rigid} objects with respect to the camera from RGB/Depth images. These methods can be divided roughly into Direct and Correspondence/Matching methods. Direct methods regress poses directly from images using deep neural networks ~\cite{li2018deepim, xiang2017posecnn} without the use of iterative Perspective-n-Point (PnP) solvers ~\cite{gao2003complete, lepetit2009ep} and they are typically formulated as a regression problem ~\cite{li2019cdpn}, ~\cite{xiang2017posecnn} or as a classification problem ~\cite{cai2022sc6d, kehl2017ssd, labbe2022megapose}. Correspondence methods ~\cite{chen2020end}, ~\cite{hodan2020epos}, ~\cite{park2019pix2pose}, ~\cite{peng2019pvnet} and feature matching methods ~\cite{fan2023pope}, ~\cite{goodwin2022zero}, ~\cite{he2022onepose++}, \cite{huang2021predator},  ~\cite{sundermeyer2023bop} use deep networks to predict 2D-to-3D correspondences between the observed image and the a 3D CAD model of the object (or 3D pointclouds of the proposals with the object surface in feature space), then predict the pose through iterative PnP solvers. Other methods ~\cite{liu2022gen6d}, ~\cite{nguyen2024gigapose}, ~\cite{cai2022ove6d} use image matching to select viewpoint rotations, followed by in-plane object rotation estimation to obtain the final estimates. \textbf{Instance-level} pose estimation methods ~\cite{ he2020pvn3d}, ~\cite{he2021ffb6d}, ~\cite{labbe2020cosypose}, ~\cite{park2019pix2pose}, ~\cite{wen2020robust} assume that \emph{exact} object 3D CAD models are available during training and testing to align a CAD model with each object observation (RGB/Depth image). \textbf{Category-level} methods ~\cite{chen2020learning}, ~\cite{lee2023tta}, ~\cite{tian2020shape}, ~\cite{wang2019normalized}, ~\cite{zhang2022ssp} handle novel object instances in a given category and don't assume that exact 3D CAD models are available for the unseen objects ~\cite{wang2019normalized}. SAM-6D ~\cite{lin2024sam} proposes a two-step category-level zero-shot 6D Pose Estimation for handling unseen objects using a semantic segmentation Vision Transformer (ViT)-based image encoder ~\cite{dosovitskiy2020image}. FoundationPose ~\cite{wen2024foundationpose} combines 6D object pose estimation and tracking, supporting both model-based and model-free setups with strong generalizability via large-scale synthetic training and language models (LLM), and transformers. 9D methods estimate the 3D rotation, translation, and additionally, the 3D dimensions of objects. They are typically applied to articulated objects such as eyeglasses and dishwashers, and rely heavily on the availability of synthetic articulated 3D pointclouds and depth maps~\cite{you2022cppf}. ~\cite{li2020category} proposes a canonical representation for different articulated objects in a given category, where the canonical object space normalizes the object orientation, scales and articulations (e.g. joint parameters and states) while each canonical part space further normalizes its part pose and scale.\\
Our method falls within the direct, regression-based, category-level family of approaches. Unlike 6D and 9D pose estimation methods, our method \emph{additionally estimates the bicycle steering and pedal rotation angles and a set of 3D bicycle Keypoints from a single bicycle image} without requiring a 3D CAD model, 3D pointcloud, nor a depth map at the input, and can work with both real and synthetic image data.


\section{Method}
\label{sec:method}
Our whole regression model architecture is illustrated in Fig. \ref{model_diagram}. Given an input image $I$ of a bicycle, our goal is to estimate the 8D bicycle pose parameters and a set of $11$ 3D bicycle Keypoints. 

\begin{figure} 
	\centering	
		\includegraphics[width=0.8\columnwidth, trim={0.8cm 18.25cm 4.0cm 0cm},clip]{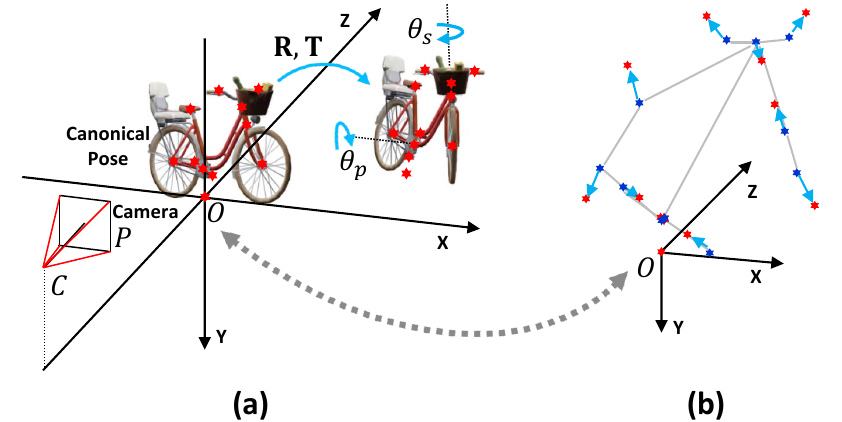} 		
	\caption{(a) Our camera setup. (b) Mean 3D Keypoints $\kappa_{\mu}$ (blue markers), and regressed 3D residual vectors $\Delta_{\kappa}$ (cyan) which point toward the red ground-truth 3D Keypoint $\bar{\kappa}_{3D}$ markers. 
	\label{camera_setup} }
\end{figure}

\textbf{Definitions:} We define the 8D pose of a 3D bicycle as the set: $\textbf{P}_{\textbf{8D}} = \{ \theta_p, \theta_s, \theta_{X}, \theta_{Y}, \theta_{Z},  \textbf{T} = (t_X$, $t_Y$, $t_Z)   \}$, where, $\theta_p$ and $\theta_s$ are the bicycle pedals axle and steering angles about their rotation shafts on the bicycle body frame (See Fig. \ref{camera_setup}), $\theta_{X}$, $\theta_{Y}$, and $\theta_{Z}$ are the bicycle body frame angles of a $3 \times 3$ rotation matrix $\textbf{R}(\theta_{X}, \theta_{Y}, \theta_{Z})$ about the world's $X$, $Y$, and $Z$ 3D axes respectively,  and $\textbf{T} = (t_X$, $t_Y$, $t_Z)$ is a 3D translation of the bicycle body with respect to the 3D origin $O=(0,0,0)$. 
We define the set of $11$ bicycle 3D Keypoints (red markers in Fig. \ref{camera_setup}) as:   $\kappa = \{$ left handle, right handle, forward wheel centre, steering axis $\#1$, steering axis $\#2$, pedal right, pedal left, pedal axle, seat (saddle), ground (root), rear wheel center$\}$.  

\textbf{Proposed Bicycle 8D Pose Regression Model}.

\textbf{Inputs.} The inputs to our model are a $512 \times 512$ RGB image $I$ and a bicycle 2D bounding box $b$ (defined on $I$), which we assume to be given by an off-the-shelf 2D object detector (2D Det) ~\cite{ge2021yolox}, ~\cite{he2017mask}. We use $I$ and $b$ to crop a sub-image $I_b$ of the bicycle, centered, upscaled to $512 \times 512$, and zero-padded. The full uncropped image $I$, which includes context, is required for accurate estimation of the 3D translation parameters ~\cite{li2022cliff}.
The sub-image $I_b$ contains important structural and appearance information from bicycle parts such as the frame, the pedals and the steering, which are key to the regression of \emph{local} bicycle rotation pose parameters such as the pedal and steering rotation angles $(\theta_p, \theta_s)$ with respect to the bicycle frame. Our regression model design has two  branches: a Translation Branch, and a Keypoint Branch, which are explained next.

\subsection{Keypoint Branch.} Given the sub-image $I_b$ and the box $b$ as inputs, the Keypoint branch regresses the bicycle body frame rotation parameters $(\theta_{X}, \theta_{Y}, \theta_{Z})$, the bicycle pedals and steering angles $(\theta_p, \theta_s)$, and a set $\Delta_{\kappa}$ of $11$ residual 3D vectors. The Keypoint Branch starts with the encoding of $I_b$ via $ViT_K$, which is an image encoder that is composed of: 1) a pre-trained vision transformer (ViT)-based image encoder (with \emph{frozen parameters}) from Segment Anything ~\cite{kirillov2023segment} (SAM ViT-b in Fig. \ref{model_diagram}), which outputs a $256 \times 64 \times 64$ latent space tensor, and 2) an adaptor $A_K$ comprised of \emph{trainable} 2D convolution stages, with batch normalization, skip paths, and ReLU activations. We re-shape the output from $ViT_K$ to a $1024$ feature vector, and concatenate it with bounding box information (following ~\cite{kanazawa2018end}) derived from $b$, including the 2D box normalized center, normalized maximum box side, and aspect ratio to get a $1028$ vector that we send to the middle-stage $L_K$, which is a set of linear layers with dropout and outputs a $256$ feature vector $f_K$. 

\textbf{Keypoint Branch Decoding Heads.} The $Dec_{2DK}$ head uses a linear layer to convert the feature vector $f_K$ into an $11$ (Keypoints) $\times$ $2$ (image coordinates) tensor that encodes the $11$ bicycle 2D Keypoint features. It then applies a self-attention stage that learns affinities between each of  these 2D Keypoint features, while weighting them via the self-attention mechanism before sending them to the final decoding linear layer with a $tanh$ activation that normalizes the $11$ regressed $I_b$ image 2D Keypoints to be within the range $[-1,1]$. The feature vector $f_K$ is also sent to: 1) $Dec_R$, which decodes the three global rotation angles $( \theta_{X}, \theta_{Y}, \theta_{Z})$, 2) $Dec_{ps}$, which decodes the pedal and steering angles $(\theta_p, \theta_s)$ in canonical bicycle pose (see Fig. \ref{camera_setup}  ), and 3) $Dec_{3D}$, which regresses 3D Keypoint residual vectors as explained next.  

\textbf{Bicycle 3D Keypoint Regression}. 
$Dec_{3D}$ (Fig. \ref{model_diagram}) decodes a set $\Delta_{\kappa}$ of $11$ canonical residual 3D vectors (cyan vectors in Fig. \ref{camera_setup}(b)). We regress canonical bicycle 3D Keypoints $\kappa_{c3D}$ by aggregating the residuals to a set $\kappa_{\mu}$ of $11$ \emph{fixed} mean canonical 3D Keypoints that we pre-computed from the training set of 3D bicycles. We compute $\kappa_{c3D}$ as shown in Eqn. \ref{eq:keypoints_mean_plus_delta}.
\begin{equation} 
\begin{aligned} 
\label{eq:keypoints_mean_plus_delta}
\kappa_{c3D} = \kappa_{\mu} + \Delta_{\kappa}.
\end{aligned}
\end{equation}

\subsection{Translation Branch} Given the image $I$ (uncropped) and the box $b$ as inputs, the Translation Branch regresses the 3D translation $\textbf{T}=(t_X, t_Y, t_Z)$ of the bicycle \emph{ground} 3D Keypoint (which is located under the pedals axle) with respect to the origin $O$ (see Fig. \ref{camera_setup}(a)). Similar to the Keypoint branch, the Translation branch also starts with the encoding of $I$ via $ViT_T$, which is an image encoder that is composed of a pre-trained (frozen) SAM ViT, and a \emph{trainable} adaptor $A_T$ stage similar to $A_K$.  We also re-shape the output from $ViT_T$ ($256 \times 2 \times 2$ tensor) to a $1024$ vector that we concatenate with bounding box information to get a $1028$ vector that we send to the middle-stage $L_T$ (similar to $L_K$) that outputs a $256$ feature vector $f_T$ that is summed with $f_K$ before it is sent to the decoding head $Dec_T$ to regress a 3D  translation vector normalized to $[-1,1]$ using a \emph{tanh} activation.  

\subsection{Parametric 3D Bicycle Model} We implemented a version of the parametric articulated bicycle model from ~\cite{corral20253darticcyclists}, which reposes (using $3D$ transformation matrices) the different bicycle parts in 3D Gaussian representation. However, our implementation operates only with 3D Keypoints and does not work with 3DGS Gaussians. The inputs to our parametric model are: 1) the regressed 8D pose parameters $\textbf{P}_{\textbf{8D}}$ and, 2) the regressed \emph{canonical} 3D Keypoints $\kappa_{c3D}$. The outputs are the \emph{updated} bicycle 3D Keypoints $\kappa_{3D}$ (i.e. the canonical Keypoints $\kappa_{c3D}$ reposed using $\textbf{P}_{\textbf{8D}}$). We use the camera matrix $P$ to project the 3D Keypoints $\kappa_{3D}$ onto the $I$ image plane, and use the bounding box $b$ information to transform the projected 2D Keypoints onto the $I_b$ image plane so that they can be used to supervise the Keypoint Branch. 


\begin{figure*} 
	\centering	
		\includegraphics[width=2.1\columnwidth, trim={0cm 10.0cm 0cm 0cm},clip]{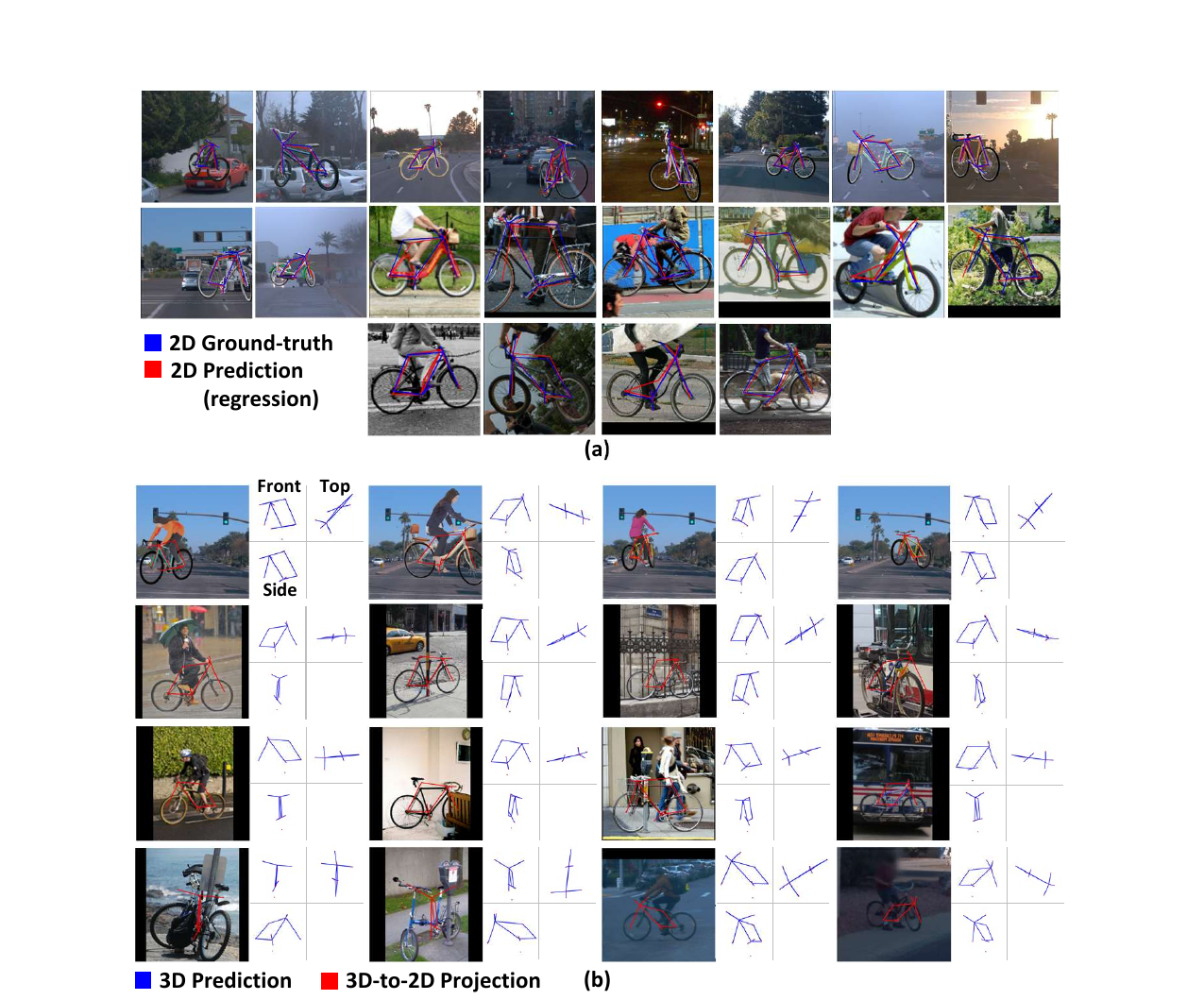} 		
	\caption{Example qualitative results. (a) 2D Keypoints regressions from our $Dec_{2DK}$ decoder. Color codes in (a) are: blue: ground-truth, red: predictions. (b) Bicycle 8D poses estimated by our full model, including the regressed (lifted) 3D bicycle Keypoints (and their 3D-to-2D projections onto the image) for both synthetic and real \emph{unseen validation frames}. Next to each image, we show three different camera views (Front, Top, Side) of the regressed (lifted) 3D Keypoints, in blue for better visualization. Typical failure cases are shown at the bottom row of (b). 
	\label{qualitative_results_fig} }
\end{figure*}

\subsection{Supervision and Training Losses}
\label{sec:training_losses}

\subsubsection{Synthetic 3D Dataset Preparation}
\label{sec:3D_ground_truth}

We need 3D bicycle training data with ground-truth 3D Keypoints to train our model, but such dataset is not available. We address this data scarcity problem by adopting the framework from ~\cite{corral20253darticcyclists} to generate synthetic 3D bicycle ground-truth (GT) data, which includes: 1) Rendered $512 \times 512$ RGB foreground bicycle images, 2) a set of ground-truth 8D bicycle pose parameters  $\overline{\textbf{P}_{\textbf{8D}}} = \{ \overline{\theta_p}, \overline{\theta_s}, \overline{\theta_{X}}, \overline{\theta_{Y}}, \overline{\theta_{Z}}, \overline{\textbf{T}}=(\overline{t_X}, \overline{t_Y}, \overline{t_Z})    \}$. We use the same camera $P$ settings from ~\cite{corral20253darticcyclists}, except, we set the camera location to $C=[0, -0.75, -12]$. We defined the following domain operating ranges for each of the ground-truth pose parameters: $\overline{\theta_p} \in  [-180^\circ, 180^\circ)$,  $\overline{\theta_s} \in  [-90^\circ,  90^\circ]$, $\overline{\theta_{X}} \in  [-5^\circ,  5^\circ]$, $\overline{\theta_{Y}} \in  [-180^\circ, 180^\circ)$, $\overline{\theta_{Z}} \in  [-5^\circ,  5^\circ]$, $\overline{t_X} \in  [-1, 1]$ m,  $\overline{t_Y} \in  [-0.5, 0.5]$ m, $\overline{t_Z} \in  [-5.0, 2.0]$ m, with respect to the origin $O=(0,0,0)$. 

The frozen SAM ViT-b image encoder from Fig. \ref{model_diagram} was pre-trained on 11 million images ~\cite{kirillov2023segment}, and does not need to be fine-tuned. To be able to train the rest of the layers from our model (convolutional/linear layers, and a single stage of self-attention), we used the camera $P$, we generated a total of $2500$ rendered synthetic images of each of the $23$ bicycles (without rider) with different 8D pose settings by performing \emph{uniform} random sampling within each ground-truth parameter domain to obtain a total of $57,500$ ground-truth bicycles. We then split (randomly) the $57,500$ bicycles into $43,125$ ($75 \%$) for training, and $14,375$ ($25 \%$) for validation. We also generated $14,400$ training and $4500$ validation frames of cyclists (i.e. bicycles \emph{with} $9$ different riders) to account for rider-related occlusions.

\subsubsection{Real 2D Dataset Preparation: 2D Keypoints}
\label{sec:2D_ground_truth}
To help our model generalize with real bicycle and cyclist images, we hand-labeled ($11$ 2D bicycle image Keypoints $\bar{\kappa}_{2D}$ and $2D$ bounding box $\bar{b}$) 
real bicycle and cyclist images from the Waymo ~\cite{sun2020scalability} and COCO $2017$ ~\cite{lin2014microsoft} datasets. We partition the real labeled data into the following sets: 1) Waymo cyclists: $216$ training frames and $72$ validation frames. 2) COCO cyclists: $198$ training frames and $67$ validation frames, and COCO bicycles (without rider): $267$ training frames, $90$ validation frames. 

\subsubsection{Training Losses}
\label{sec:training_losses}
To train our bicycle pose estimation model we defined the following seven loss terms (See red boxes in Fig. \ref{model_diagram}): 1) 3D translation loss $\mathcal{L}_{T}(\textbf{T}, \bar{\textbf{T}})$, 2) 3D bicycle body frame rotation angle loss $\mathcal{L}_{R}(\textbf{R}, \bar{\textbf{R}})$, 3) Bicycle pedals and steering rotation angle loss $\mathcal{L}_{ps}((\theta_p, \theta_s), (\bar{\theta_p}, \bar{\theta_s}))$, 4) 3D Keypoint residual loss $\mathcal{L}_{3D}( \Delta_{\kappa_{3D}} , \overline{\Delta_{\kappa_{3D}} } )$, 5) 2D Keypoint loss $\mathcal{L}_{2DK}(\kappa_{2D}, \overline{\kappa_{2D}} )$, and 6) 2D Keypoint consistency loss $\mathcal{L}_{2Dcon}(\kappa_{2D}, H_b(Proj(\kappa_{3D}, P)) )$, where $Proj$ denotes a 3D-to-2D projection using camera $P$, and $H_b$ is a function that maps 2D Keypoints defined on the image $I$ onto the image $I_b$. This loss term encourages consistency between regressed 2D Keypoints and the projections onto the image plane of the regressed 3D Keypoints. Finally, 7) $L_{aux}$, is an auxiliary loss that will be explained in the next sub-section. All losses are $L_2$, and they are computed using regressed and ground-truth quantities normalized with respect to pre-defined domain ranges for each parameter as explained in Section \ref{sec:3D_ground_truth}.

The total loss $\mathcal{L}_{total}$ is:
\begin{equation} 
\begin{aligned} 
\label{eq:loss_bike_total}
\mathcal{L}_{total} = & \beta_1 \mathcal{L}_{R} + \beta_2 \mathcal{L}_{T} + \beta_3 \mathcal{L}_{ps} + \beta_4 \mathcal{L}_{3D} + \\
                                   & \beta_5 \mathcal{L}_{2DK} + \beta_6 \mathcal{L}_{2Dcon} + \beta_7 \mathcal{L}_{aux} ,
\end{aligned}
\end{equation}
where $\beta_1=1$, $\beta_2=1$, $\beta_3=2$, $\beta_4=1/2$, $\beta_5=1$, $\beta_6=1$, and $\beta_7=0.2$ are scalars to control the contribution from each loss term, and their values were determined experimentally.

\subsubsection{Training Strategy}
\label{sec:training_strategy}
We train our model end-to-end by inputting one batch of real data (2D) into the model for every $N$ batches of synthetic data (3D), where we $N=10$ experimentally. 

\textbf{Training with Synthetic Data Batches}. We overlay each input image $I$ (foreground synthetic bicycle image with white background), onto a randomly-selected real Waymo ~\cite{sun2020scalability} background non-bicycle image via alpha blending (similar to ~\cite{zuffi2018lions}). We un-freeze all of the model layers (Translation and Keypoint branches) so that all model parameters are updated using Eqn. \ref{eq:loss_bike_total}. 

\textbf{Training with Real Data Batches (2D)}. For the real data case, since there is no 3D ground-truth, we can only train the $A_K$ ($ViT_K$) , $L_K$ and $Dec_{2DK}$ layers from the Keypoint branch. To be able to train Translation branch layers with real images, we implemented a simple MLP-based auxiliary task ~\cite{corral2023domain} decoder $Dec_{aux}$ (shown in Fig.  \ref{model_diagram}) that regresses the center of the 2D bounding box. Having $Dec_{aux}$ enables us to train (at least) the $A_T$ convolutional stages from the $ViT_T$ encoder with real images, which is beneficial to improve generalization with real bicycle images. We freeze all model layers except $A_T$, $Dec_{aux}$, $A_K$, $L_K$ and $Dec_{2DK}$, effectively setting $\mathcal{L}_{total} = \beta_5 \mathcal{L}_{2DK} + \beta_7 \mathcal{L}_{aux} $. 

\textbf{Data Augmentations}.
When training with synthetic data batches, we add appearance perturbations to the foreground bicycle/cyclist image (without altering the background image), by randomly rotating the UV color plane in YUV color space by $+/- 30^\circ$. with probability $p=0.25$. When training with real data batches we perform random image horizontal flip with probability $p=0.5$. 


\begin{table*}
	\caption{Per-Pose Parameter Mean Absolute Error (\textbf{MAE}) $\downarrow$ with respect to ground-truth. The rotation MAEs are in degrees, the translation MAEs are in meters. Both, SAM-6D ~\cite{lin2024sam} and Foundationpose ~\cite{wen2024foundationpose} require a dense 3D pointcloud, depth map, and a corresponding RGB image as inputs. Our method only expects an RGB image and a 2D bounding box $b$ as inputs. Evaluation performed only on \emph{unseen} synthetic images from the \emph{validation} set.}  
	\label{eval_table_params_mae}
	\centering
       \begin{tabular}{lllllllllll}	 
       \toprule
	        Method               & $\theta_{X}^{MAE}$    & $\theta_{Y}^{MAE}$   & $\theta_{Z}^{MAE}$  & $t_X^{MAE}$               & $t_Y^{MAE}$          & $t_Z^{MAE}$          & $\theta_{p}^{MAE}$  &  $\theta_{s}^{MAE}$  \\
	 \hline
            Foundationpose ~\cite{wen2024foundationpose}      & 8.31            & 18.89          & 7.41         & 0.063    & 0.066   & 0.257      &               &      \\
            \hline
            SAM-6D ~\cite{lin2024sam}       & 5.85        & 13.41          & 9.29     & 0.065   & 0.076   & \textbf{0.086}                &               &      \\
            SAM-6D FT                               & 9.19        & 53.07          & 28.89     & 0.144   & 0.264   & 0.139                &               &      \\
            \hline             
            \textbf{Ours} (RGB only)   & \textbf{1.91}   & \textbf{6.02}    & \textbf{1.24}   & \textbf{0.022}    & \textbf{0.020}  & 0.251       &  \textbf{30.98}        & \textbf{25.27} \\	

	      \bottomrule
	\end{tabular}
\end{table*}

\section{Experiments and results}
\label{sec:experiments}

\begin{table*}
	\caption{3D Average Recall (AR) and other 6D evaluation metrics computed only on \emph{unseen} synthetic images from the \emph{validation} set.} 
	\label{eval_table_ar_add}
	\centering
       \begin{tabular}{lllllllllll}	 
        \toprule 
	      Method                      & $3D_{10}$$\uparrow$   & $3D_{25}$$\uparrow$   &  $3D_{50}$$\uparrow$  & 5$^\circ$,5cm$\uparrow$ & 10$^\circ$,10cm$\uparrow$  & 40$^\circ$,20cm$\uparrow$ & 60$^\circ$,30cm$\uparrow$  &  ADD$\downarrow$  \\
	 \hline
            Foundationpose ~\cite{wen2024foundationpose}    & 95.84       &  89.69       & 78.53            & 3.57                 & 18.55                      & 60.26                       &  79.67              & 0.196           \\
            \hline
           SAM-6D ~\cite{lin2024sam}      & 97.37   &  \textbf{95.11}   & \textbf{91.42}       & \textbf{29.21}     & \textbf{78.22}     & \textbf{91.05}       & \textbf{92.24}      & \textbf{0.105}       \\ 
           SAM-6D FT                              & \textbf{99.52}   &  92.89   & 77.92       &  0.00     &   9.06     & 51.39       &  58.91      & 0.319       \\ 
            \hline          
            \textbf{Ours} (RGB only)                  & 99.20           & 91.50     & 59.40          & 5.03                  & 20.42                   & 45.23                      &  64.41             &  0.271       \\

           \bottomrule

	\end{tabular}
\end{table*}

\begin{table*}
	\caption{2D Keypoint Average Recall (AR) on $512 \times 512$ pixel images from different \emph{unseen validation} image sets.} 
	\label{eval_table_ar_2D}
	\centering
	\begin{tabular}{llllllll}
		\toprule
		Exp. &Validation Dataset   & 2D Keypoints  & $2D_{5pxl}$$\uparrow$   & $2D_{10pxl}$$\uparrow$   &  $2D_{20pxl}$$\uparrow$  & $2D_{30pxl}$$\uparrow$        \\
		\midrule
		1 &  COCO (real)         & $Proj(\kappa_{3D}, P)$                                               & 3.14   & 10.62  & 31.92  & 50.05                   \\ 
             2 & Waymo (real)      & $Proj(\kappa_{3D}, P)$                                               & 2.86   & 8.70    & 25.88  & 43.41                \\ 
             3 & 3DArticCyclists (Synthetic) & $Proj(\kappa_{3D}, P)$ & 7.68   & 23.72  & 52.76  & 70.53                \\ 
		4 & 3DArticCyclists (Synthetic)  & $H_b(Proj(\kappa_{3D}, P))$                                     & \textbf{22.53}  & \textbf{48.74}  & \textbf{78.77}  & \textbf{91.33}                             \\
		\bottomrule
	\end{tabular}
\end{table*}

\begin{table*}
	\caption{Ablation experiments. Evaluations performed on the \emph{validation set}.} 
	\label{eval_table_ablations}
	\centering
	\begin{tabular}{lllllllllllllll}
		\toprule
		    & $\mathcal{L}_{R}$      & $\mathcal{L}_{T}$     & $\mathcal{L}_{ps}$  & $\mathcal{L}_{3D}$     & $\mathcal{L}_{2DK}$ & $\mathcal{L}_{2Dcon}$     & $\mathcal{L}_{aux}$     & $3D_{25}$  & $3D_{50}$ & $5^\circ,5cm$    &  ADD &  $\theta_{p}^{MAE}$ &  $\theta_{s}^{MAE}$     \\
		\midrule
		1 (8D)  & \checkmark    &  \checkmark  & \checkmark  & \checkmark  & \checkmark  & \checkmark  & \checkmark  & 91.50  & 59.40  & 5.03  &  0.271  & 30.98  & 25.27  \\
                     2  & \checkmark    &  \checkmark  &                   & \checkmark  & \checkmark  & \checkmark   & \checkmark & 88.69  & 50.83  & 4.82    &  0.305  &              &             \\
		3  & \checkmark    &  \checkmark  &                   &                   & \checkmark  & \checkmark   & \checkmark  & 85.80  & 46.21 & 5.53      &  0.325 &              &             \\
		4  & \checkmark    &  \checkmark  & \checkmark & \checkmark  &                    &                    & \checkmark  & 87.20  & 49.62 & 3.96      &  0.307  & 31.12  & 24.60  \\
                     5 (6D) & \checkmark    &  \checkmark  &                   &                   &                      &                    &                     & 96.95  & 73.20 & 7.51      &  0.226  &              &             \\
		\bottomrule
	\end{tabular}
\end{table*}

We performed two types of evaluations of our 8D bicycle pose estimation model. First, we evaluated the accuracy of each of the individual estimated pose parameters. Then, we followed evaluation protocols that are commonly used in 6D pose estimation methods, while also evaluating the quality of our regressed 3D Keypoints and their 2D projections onto the image plane. We perform all of our evaluations on \emph{unseen frames from the validation sets}.

\textbf{Baseline Methods}. We could not find any 8D baseline method that estimates the two new steering and pedal angles from articulated bicycles. Nevertheless, we can still compare the 6D pose estimation part of our method against recent state-of-the art 6D pose estimation methods. For this, we chose the SAM-6D ~\cite{lin2024sam} and Foundationpose ~\cite{wen2024foundationpose} methods from robotics as 6D baselines. These methods can estimate the 6D pose (i.e. rotation $R$, and translation $T$) of novel rigid objects, without retraining (\cite{lin2024sam}, \cite{wen2024foundationpose}), as long as they are provided with: an RGB image, a CAD 3D model of the new object, a metric depth map, and camera intrinsic parameters. Foundationpose only provides code/guidelines for inference, but not for training. SAM-6D also provides code/gudelines mainly for inference, however, a training script can be found in their code repository, but it comes  without instructions/guidelines. We adapted these two methods to work with our synthetic articulated bicycles and cyclist data. For each frame, we prepared the required dense 3D mesh/pointcloud and depth map. For a fair comparison (since we assume that a 2D bounding box is given to our method), we also prepared a binary mask of the cyclist for each input image to pre-segment the input object in these methods (i.e. to avoid SAM segmentation failures and to fully focus on the pose estimation problem). First, we ran both of the SAM-6D and Foundationpose 6D methods in inference mode, using rigid templates with $\theta_p=\theta_s=0$. For all of the template/hypothesis 3D pointclouds, the bicycle body frame rotation $R$ is set to the identity matrix, and the 3D location is set to $T=(0,0,0)$. We used their default configuration settings, except we disabled the pose tracking in Foundationpose because our data is not sequential.
Then, for SAM-6D, we used their training script without any changes to run a finetuning (8 epochs) with our data using their default settings, but experimenting with different/lower learning rates. Although we observed the loss being minimized during training, we noticed  that the inference is very sensitive to the number of 3D templates, and had to set the number of templates to $2$ (the training script uses two templates) to obtain reasonable results.

\subsection{Pose Parameter Evaluation using MAE}
\label{sec:pose_params_evaluation}

We ran both, our trained model and the baselines on \emph{unseen} images (\emph{from the validation set}) of cyclists overlaid on a real background, and computed the Mean Absolute Error (MAE) for each of the independent estimated parameters with respect to their ground-truth. For angles, we computed the MAE using the smaller angular differences between estimated and ground-truth. Table \ref{eval_table_params_mae} shows the results. Our RGB-only method has the lowest mean absolute error for the global rotation parameters $\theta_X$, $\theta_y$, $\theta_Z$, and for the translation parameters $t_X$ and $t_Y$. We believe that the higher $t_Z=0.251$ MAE relates to the lack of use of depth maps in our method. Moreover, \textbf{ours is the only method that can estimate the pedal and steering angles} ($\theta_p$, $\theta_s$), with mean absolute errors of $30.98^\circ$ and $25.27^\circ$ respectively. These errors are relatively higher that those from the global rotation angles, which shows how \emph{challenging} is to estimate these parameters from a single image visually: the bicycle pedals and steering parts are relatively small and thin compared to the whole bicycle body frame part. And they are often occluded on the observed image. Nevertheless, being able to obtain estimates of these two new parameters can be useful for pose-based start intention detection of cyclists based on the pedals angle visual cue  ~\cite{kress2019pose, zhang2013rider}, for human-object interaction applications ~\cite{yang2024lemon}, and for determining a more fine-grained bicycle pose state and a better estimated travel direction as shown in Fig. \ref{Cano_vs_Posed_Mismatch}.

\subsection{Evaluation Using 6D Pose Estimation Metrics} 
\label{sec:pose_evaluation_metrics} 
We adopt the Average Recall (AR) metric used in 6D pose estimation methods ~\cite{hodavn2016evaluation, labbe2022megapose, wang2019normalized, wen2024foundationpose} using different criteria such as 3D bounding box Intersection Over the Union (IoU) and the ($\alpha^\circ$, $\delta$ m) pose rotational and translational error metric which considers an estimated pose to be correct if its rotation error is within $\alpha^\circ$ and the translation error is below $\delta$ m   ~\cite{li2018deepim}. We computed the rotational and translational errors as explained in ~\cite{hodavn2016evaluation}. We defined ground-truth 3D bounding boxes derived from the canonical bicycle 3D model dimensions, and from the GT 3D Keypoints. Initially, these boxes are in canonical pose (rotation $R$ set to the identity matrix, and the 3D location is set to $T=(0,0,0)$), then we repose them using both, the GT and the estimated 6D poses from each method to compute the evaluation metrics using the 3D box IoU implementation from Objectron ~\cite{ahmadyan2021objectron}. To evaluate the quality of the regressed 3D Keypoints, we implemented the Average Distance of Model Points (ADD) in meters, as described in ~\cite{xiang2017posecnn} and ~\cite{hodavn2016evaluation}. Table \ref{eval_table_ar_add} summarizes the results, where SAM-6D is the clear winner, however, \textbf{our RGB-only method achieves competitive scores} for $3D_{10}$ ($99.20$) and $3D_{25}$ ($91.50$), and outperformed Foundationpose in those metrics. Our method also outperformed Foundationpose in the $5^\circ, 5cm $ and $10^\circ, 10cm$ criteria, which is encouraging. We believe that both, SAM-6D and Foundationpose perform better than our method because they have access to metric depth maps and 3D pointclouds in inference mode, which enables them to obtain an accurate estimate of the 3D translation $T$ and perform iterative refinement of the pose ~\cite{xiang2017posecnn}.

\subsection{2D Keypoint Evaluation}
\label{sec:2D_evaluation}

We also evaluated the accuracy of our method's regressed 2D Keypoints on the image plane using a 2D Average Recall (AR) metric based on 2D pixel distances using four different criteria: $2D_{5pxl}$, $2D_{10pxl}$, $2D_{20pxl}$, and $2D_{30pxl}$. We ran our method separately on the \emph{validation} images from each of the COCO, Waymo, and 3DArticCyclists datasets. Table \ref{eval_table_ar_2D} summarizes the results. Rows 1, 2, and 3 from the table show results for 2D Keypoints defined on the original image $I$. Row 4 is for 2D Keypoints defined on the cropped sub-image $I_b$. For the first three rows we observe that the performance with COCO images is higher than with Waymo images. We attribute this to both, the number of training images, and the quality of the images (Waymo's cyclist images are usually blurred with poor contrast).  Similarly, the performance with 3DArticCyclists images is higher than with COCO images, likely due to the higher number and quality of the synthetic images. On the other hand, the performance in Row 4 is clearly higher than that from Row 3, which is surprising. We theorize that this happens because:  1) The bicycle's cropped sub-image image $I_b$ has a larger effective receptive field when processed by our model, and 2) The 2D Keypoints defined on $I_b$ are the actual supervision signal used to train $Dec_{2DK}$, including the self-attention stage from the Keypoint branch. In general, we believe that having significantly larger labeled real image sets for training would help improve generalization with real images.

\subsection{Qualitative Results}
\label{sec:qualitative_results}

Fig. \ref{qualitative_results_fig} shows example qualitative results from our 8D bicycle pose estimator, on both, synthetic and real bicycles with and without rider. Fig. \ref{qualitative_results_fig}(a) shows 2D Keypoint regressions from our $Dec_{2DK}$ decoder. Fig. \ref{qualitative_results_fig}(b) shows visualizations of our estimated bicycle 8D poses, where next to each image, we show three different views (Front, Top, Side) of the regressed (lifted) 3D Keypoints. We also include examples of the typical failure cases that we observe in our method, shown at the bottom row of the figure.

\subsection{Ablation studies}
\label{sec:ablation}
We are interested in understanding the contribution from key stages of our model. Specifically, we focus on 1) $Dec_{ps}$ that regresses the new pedals and steering angles, 2) $Dec_{3D}$, which regresses the 3D residual vectors used to regress the 3D Keypoints, and 3) $Dec_{2DK}$ that regresses the 2D Keypoints using self-attention to learn the features $f_K$. Table \ref{eval_table_ablations} summarizes our training experiments, where we disabled some of these stages via setting their corresponding losses to zero. We perform the evaluation using metrics from Tables \ref{eval_table_ar_add} (AR, ADD) and \ref{eval_table_params_mae} ($\theta_p$ and $\theta_s$ MAE). Table \ref{eval_table_ablations} Row 1 shows results from our full 8D model. Row 2 shows degradations of disabling the regression of the new parameters $\theta_p$ and $\theta_s$. In Row 3 we further disable the regression of $\Delta_{K}$. Row 4 shows the effects of disabling the 2D  Keypoint self-attention-based $Dec_{2DK}$ stage alone. Row 5 shows our method in 6D pose estimation mode, which we can interpret as a \textbf{vanilla RGB-only 6D pose estimation method}.  This vanilla method is trained to exclusively regress the 3D rotation and translation parameters, and it performs better on the 6D evaluation metrics, but it looses the ability to regress $\theta_p$, $\theta_s$, and the set $\kappa_{3D}$ of 3D bicycle Keypoints.

\subsection{Training details}
\label{sec:3D_bike_pose_model_training_details}

We train our model using the Adam optimizer for $20$ epochs, with learning rate $LR = 0.001$, using a batch size of $10$, on a single  
NVidia Tesla V100 GPU with 32510 MiB.  In inference mode it processes one frame in $0.332$ seconds. 

\subsection{Discussion and Limitations}
\label{sec:limitations}

Our 8D pose estimation method has been tested only with bicycles and cyclists. In our method, the lack of metric depth maps for refinement results in higher 3D translation ($t_X$, $t_Y$, $t_Z$) errors, which in turn cause 3D global shifts of the estimated 3D Keypoints. The effects are visible: 1) in the ADD metric (higher distance error between predicted and GT 3D points due to 3D shift), and 2) in the 3D-to-2D projected Keypoints of real images (see real images from Fig. \ref{qualitative_results_fig}-b, where the bicycle local pose is correct, but not well aligned to the bicycle on the image because of the 3D shift). A potential solution for the case of the synthetic data would be to introduce metric depth, however \emph{metric depth maps are not available for the real COCO/Waymo images}. Obtaining accurate estimates of the two new parameters from a single image is a very challenging problem (these parts are relatively small and often occluded on the image). Using auxiliary information such as symmetry could be exploited to obtain better estimates.

\section{Conclusions}
\label{sec:conclusions}

We have a presented a new monocular visual 8D pose estimation method for articulated bicycles and cyclists that can estimate the pedal and steering angles, besides the 6D rotation and translation parameters and the bicycle 3D Keypoints from a single RGB image, without additional inputs such as depth maps, point clouds, or CAD models. We believe that our method can be extended for both indoor and outdoor applications such as robotic manipulation of articulated objects and cyclist road crossing intention prediction in autonomous driving.



\bibliographystyle{IEEEtran}
\bibliography{egbib_cyclist_arxiv_2025}

\begin{thebibliography}{10}
\providecommand{\url}[1]{#1}
\csname url@rmstyle\endcsname
\providecommand{\newblock}{\relax}
\providecommand{\bibinfo}[2]{#2}
\providecommand\BIBentrySTDinterwordspacing{\spaceskip=0pt\relax}
\providecommand\BIBentryALTinterwordstretchfactor{4}
\providecommand\BIBentryALTinterwordspacing{\spaceskip=\fontdimen2\font plus
\BIBentryALTinterwordstretchfactor\fontdimen3\font minus
  \fontdimen4\font\relax}
\providecommand\BIBforeignlanguage[2]{{%
\expandafter\ifx\csname l@#1\endcsname\relax
\typeout{** WARNING: IEEEtran.bst: No hyphenation pattern has been}%
\typeout{** loaded for the language `#1'. Using the pattern for}%
\typeout{** the default language instead.}%
\else
\language=\csname l@#1\endcsname
\fi
#2}}

\bibitem{abadi2023detection}
A.~D. Abadi, Y.~Gu, I.~Goncharenko, and S.~Kamijo, ``Detection of cyclist’s
  crossing intention based on posture estimation for autonomous driving,''
  \emph{IEEE Sensors Journal}, vol.~23, no.~11, pp. 11\,274--11\,284, 2023.

\bibitem{kress2019pose}
V.~Kress, J.~Jung, S.~Zernetsch, K.~Doll, and B.~Sick, ``Pose based start
  intention detection of cyclists,'' in \emph{2019 IEEE Intelligent
  Transportation Systems Conference (ITSC)}.\hskip 1em plus 0.5em minus
  0.4em\relax IEEE, 2019, pp. 2381--2386.

\bibitem{gu2014recognition}
Y.~Gu and S.~Kamijo, ``Recognition and pose estimation of urban road users from
  on-board camera for collision avoidance,'' in \emph{17th International IEEE
  Conference on Intelligent Transportation Systems (ITSC)}.\hskip 1em plus
  0.5em minus 0.4em\relax IEEE, 2014, pp. 1266--1273.

\bibitem{cho2010vision}
H.~Cho, P.~E. Rybski, and W.~Zhang, ``Vision-based bicycle detection and
  tracking using a deformable part model and an ekf algorithm,'' in \emph{13th
  International IEEE Conference on Intelligent Transportation Systems}.\hskip
  1em plus 0.5em minus 0.4em\relax IEEE, 2010, pp. 1875--1880.

\bibitem{lin2024sam}
J.~Lin, L.~Liu, D.~Lu, and K.~Jia, ``Sam-6d: Segment anything model meets
  zero-shot 6d object pose estimation,'' in \emph{Proceedings of the IEEE/CVF
  Conference on Computer Vision and Pattern Recognition}, 2024, pp.
  27\,906--27\,916.

\bibitem{wen2024foundationpose}
B.~Wen, W.~Yang, J.~Kautz, and S.~Birchfield, ``Foundationpose: Unified 6d pose
  estimation and tracking of novel objects,'' in \emph{Proceedings of the
  IEEE/CVF Conference on Computer Vision and Pattern Recognition}, 2024, pp.
  17\,868--17\,879.

\bibitem{labbe2022megapose}
Y.~Labb{\'e}, L.~Manuelli, A.~Mousavian, S.~Tyree, S.~Birchfield, J.~Tremblay,
  J.~Carpentier, M.~Aubry, D.~Fox, and J.~Sivic, ``Megapose: 6d pose estimation
  of novel objects via render \& compare,'' \emph{arXiv preprint
  arXiv:2212.06870}, 2022.

\bibitem{li2018deepim}
Y.~Li, G.~Wang, X.~Ji, Y.~Xiang, and D.~Fox, ``Deepim: Deep iterative matching
  for 6d pose estimation,'' in \emph{Proceedings of the European Conference on
  Computer Vision (ECCV)}, 2018, pp. 683--698.

\bibitem{xiang2017posecnn}
Y.~Xiang, T.~Schmidt, V.~Narayanan, and D.~Fox, ``Posecnn: A convolutional
  neural network for 6d object pose estimation in cluttered scenes,''
  \emph{arXiv preprint arXiv:1711.00199}, 2017.

\bibitem{gao2003complete}
X.-S. Gao, X.-R. Hou, J.~Tang, and H.-F. Cheng, ``Complete solution
  classification for the perspective-three-point problem,'' \emph{IEEE
  transactions on pattern analysis and machine intelligence}, vol.~25, no.~8,
  pp. 930--943, 2003.

\bibitem{lepetit2009ep}
V.~Lepetit, F.~Moreno-Noguer, and P.~Fua, ``Ep n p: An accurate o (n) solution
  to the p n p problem,'' \emph{International journal of computer vision},
  vol.~81, pp. 155--166, 2009.

\bibitem{li2019cdpn}
Z.~Li, G.~Wang, and X.~Ji, ``Cdpn: Coordinates-based disentangled pose network
  for real-time rgb-based 6-dof object pose estimation,'' in \emph{Proceedings
  of the IEEE/CVF international conference on computer vision}, 2019, pp.
  7678--7687.

\bibitem{cai2022sc6d}
D.~Cai, J.~Heikkil{\"a}, and E.~Rahtu, ``Sc6d: Symmetry-agnostic and
  correspondence-free 6d object pose estimation,'' in \emph{2022 International
  Conference on 3D Vision (3DV)}.\hskip 1em plus 0.5em minus 0.4em\relax IEEE,
  2022, pp. 536--546.

\bibitem{kehl2017ssd}
W.~Kehl, F.~Manhardt, F.~Tombari, S.~Ilic, and N.~Navab, ``Ssd-6d: Making
  rgb-based 3d detection and 6d pose estimation great again,'' in
  \emph{Proceedings of the IEEE international conference on computer vision},
  2017, pp. 1521--1529.

\bibitem{chen2020end}
B.~Chen, A.~Parra, J.~Cao, N.~Li, and T.-J. Chin, ``End-to-end learnable
  geometric vision by backpropagating pnp optimization,'' in \emph{Proceedings
  of the IEEE/CVF Conference on Computer Vision and Pattern Recognition}, 2020,
  pp. 8100--8109.

\bibitem{hodan2020epos}
T.~Hodan, D.~Barath, and J.~Matas, ``Epos: Estimating 6d pose of objects with
  symmetries,'' in \emph{Proceedings of the IEEE/CVF conference on computer
  vision and pattern recognition}, 2020, pp. 11\,703--11\,712.

\bibitem{park2019pix2pose}
K.~Park, T.~Patten, and M.~Vincze, ``Pix2pose: Pixel-wise coordinate regression
  of objects for 6d pose estimation,'' in \emph{Proceedings of the IEEE/CVF
  international conference on computer vision}, 2019, pp. 7668--7677.

\bibitem{peng2019pvnet}
S.~Peng, Y.~Liu, Q.~Huang, X.~Zhou, and H.~Bao, ``Pvnet: Pixel-wise voting
  network for 6dof pose estimation,'' in \emph{Proceedings of the IEEE/CVF
  conference on computer vision and pattern recognition}, 2019, pp. 4561--4570.

\bibitem{fan2023pope}
Z.~Fan, P.~Pan, P.~Wang, Y.~Jiang, D.~Xu, H.~Jiang, and Z.~Wang, ``Pope: 6-dof
  promptable pose estimation of any object,'' \emph{Any Scene, with One
  Reference. arXiv}, vol. 2305, 2023.

\bibitem{goodwin2022zero}
W.~Goodwin, S.~Vaze, I.~Havoutis, and I.~Posner, ``Zero-shot category-level
  object pose estimation,'' in \emph{European Conference on Computer
  Vision}.\hskip 1em plus 0.5em minus 0.4em\relax Springer, 2022, pp. 516--532.

\bibitem{he2022onepose++}
X.~He, J.~Sun, Y.~Wang, D.~Huang, H.~Bao, and X.~Zhou, ``Onepose++:
  Keypoint-free one-shot object pose estimation without cad models,''
  \emph{Advances in Neural Information Processing Systems}, vol.~35, pp.
  35\,103--35\,115, 2022.

\bibitem{huang2021predator}
S.~Huang, Z.~Gojcic, M.~Usvyatsov, A.~Wieser, and K.~Schindler, ``Predator:
  Registration of 3d point clouds with low overlap,'' in \emph{Proceedings of
  the IEEE/CVF Conference on computer vision and pattern recognition}, 2021,
  pp. 4267--4276.

\bibitem{sundermeyer2023bop}
M.~Sundermeyer, T.~Hoda{\v{n}}, Y.~Labbe, G.~Wang, E.~Brachmann, B.~Drost,
  C.~Rother, and J.~Matas, ``Bop challenge 2022 on detection, segmentation and
  pose estimation of specific rigid objects,'' in \emph{Proceedings of the
  IEEE/CVF Conference on Computer Vision and Pattern Recognition}, 2023, pp.
  2785--2794.

\bibitem{liu2022gen6d}
Y.~Liu, Y.~Wen, S.~Peng, C.~Lin, X.~Long, T.~Komura, and W.~Wang, ``Gen6d:
  Generalizable model-free 6-dof object pose estimation from rgb images,'' in
  \emph{European Conference on Computer Vision}.\hskip 1em plus 0.5em minus
  0.4em\relax Springer, 2022, pp. 298--315.

\bibitem{nguyen2024gigapose}
V.~N. Nguyen, T.~Groueix, M.~Salzmann, and V.~Lepetit, ``Gigapose: Fast and
  robust novel object pose estimation via one correspondence,'' in
  \emph{Proceedings of the IEEE/CVF Conference on Computer Vision and Pattern
  Recognition}, 2024, pp. 9903--9913.

\bibitem{cai2022ove6d}
D.~Cai, J.~Heikkil{\"a}, and E.~Rahtu, ``Ove6d: Object viewpoint encoding for
  depth-based 6d object pose estimation,'' in \emph{Proceedings of the IEEE/CVF
  Conference on Computer Vision and Pattern Recognition}, 2022, pp. 6803--6813.

\bibitem{he2020pvn3d}
Y.~He, W.~Sun, H.~Huang, J.~Liu, H.~Fan, and J.~Sun, ``Pvn3d: A deep point-wise
  3d keypoints voting network for 6dof pose estimation,'' in \emph{Proceedings
  of the IEEE/CVF conference on computer vision and pattern recognition}, 2020,
  pp. 11\,632--11\,641.

\bibitem{he2021ffb6d}
Y.~He, H.~Huang, H.~Fan, Q.~Chen, and J.~Sun, ``Ffb6d: A full flow
  bidirectional fusion network for 6d pose estimation,'' in \emph{Proceedings
  of the IEEE/CVF conference on computer vision and pattern recognition}, 2021,
  pp. 3003--3013.

\bibitem{labbe2020cosypose}
Y.~Labb{\'e}, J.~Carpentier, M.~Aubry, and J.~Sivic, ``Cosypose: Consistent
  multi-view multi-object 6d pose estimation,'' in \emph{Computer Vision--ECCV
  2020: 16th European Conference, Glasgow, UK, August 23--28, 2020,
  Proceedings, Part XVII 16}.\hskip 1em plus 0.5em minus 0.4em\relax Springer,
  2020, pp. 574--591.

\bibitem{wen2020robust}
B.~Wen, C.~Mitash, S.~Soorian, A.~Kimmel, A.~Sintov, and K.~E. Bekris,
  ``Robust, occlusion-aware pose estimation for objects grasped by adaptive
  hands,'' in \emph{2020 IEEE International Conference on Robotics and
  Automation (ICRA)}.\hskip 1em plus 0.5em minus 0.4em\relax IEEE, 2020, pp.
  6210--6217.

\bibitem{chen2020learning}
D.~Chen, J.~Li, Z.~Wang, and K.~Xu, ``Learning canonical shape space for
  category-level 6d object pose and size estimation,'' in \emph{Proceedings of
  the IEEE/CVF conference on computer vision and pattern recognition}, 2020,
  pp. 11\,973--11\,982.

\bibitem{lee2023tta}
T.~Lee, J.~Tremblay, V.~Blukis, B.~Wen, B.-U. Lee, I.~Shin, S.~Birchfield,
  I.~S. Kweon, and K.-J. Yoon, ``Tta-cope: Test-time adaptation for
  category-level object pose estimation,'' in \emph{Proceedings of the IEEE/CVF
  Conference on Computer Vision and Pattern Recognition}, 2023, pp.
  21\,285--21\,295.

\bibitem{tian2020shape}
M.~Tian, M.~H. Ang, and G.~H. Lee, ``Shape prior deformation for categorical 6d
  object pose and size estimation,'' in \emph{Computer Vision--ECCV 2020: 16th
  European Conference, Glasgow, UK, August 23--28, 2020, Proceedings, Part XXI
  16}.\hskip 1em plus 0.5em minus 0.4em\relax Springer, 2020, pp. 530--546.

\bibitem{wang2019normalized}
H.~Wang, S.~Sridhar, J.~Huang, J.~Valentin, S.~Song, and L.~J. Guibas,
  ``Normalized object coordinate space for category-level 6d object pose and
  size estimation,'' in \emph{Proceedings of the IEEE/CVF Conference on
  Computer Vision and Pattern Recognition}, 2019, pp. 2642--2651.

\bibitem{zhang2022ssp}
R.~Zhang, Y.~Di, F.~Manhardt, F.~Tombari, and X.~Ji, ``Ssp-pose: Symmetry-aware
  shape prior deformation for direct category-level object pose estimation,''
  in \emph{2022 IEEE/RSJ International Conference on Intelligent Robots and
  Systems (IROS)}.\hskip 1em plus 0.5em minus 0.4em\relax IEEE, 2022, pp.
  7452--7459.

\bibitem{dosovitskiy2020image}
A.~Dosovitskiy, ``An image is worth 16x16 words: Transformers for image
  recognition at scale,'' \emph{arXiv preprint arXiv:2010.11929}, 2020.

\bibitem{you2022cppf}
Y.~You, R.~Shi, W.~Wang, and C.~Lu, ``Cppf: Towards robust category-level 9d
  pose estimation in the wild,'' in \emph{Proceedings of the IEEE/CVF
  Conference on Computer Vision and Pattern Recognition}, 2022, pp. 6866--6875.

\bibitem{li2020category}
X.~Li, H.~Wang, L.~Yi, L.~J. Guibas, A.~L. Abbott, and S.~Song,
  ``Category-level articulated object pose estimation,'' in \emph{Proceedings
  of the IEEE/CVF conference on computer vision and pattern recognition}, 2020,
  pp. 3706--3715.

\bibitem{ge2021yolox}
Z.~Ge, S.~Liu, F.~Wang, Z.~Li, and J.~Sun, ``Yolox: Exceeding yolo series in
  2021,'' \emph{arXiv preprint arXiv:2107.08430}, 2021.

\bibitem{he2017mask}
K.~He, G.~Gkioxari, P.~Doll{\'a}r, and R.~Girshick, ``Mask r-cnn,'' in
  \emph{Proceedings of the IEEE international conference on computer vision},
  2017, pp. 2961--2969.

\bibitem{li2022cliff}
Z.~Li, J.~Liu, Z.~Zhang, S.~Xu, and Y.~Yan, ``Cliff: Carrying location
  information in full frames into human pose and shape estimation,'' in
  \emph{European Conference on Computer Vision}.\hskip 1em plus 0.5em minus
  0.4em\relax Springer, 2022, pp. 590--606.

\bibitem{kirillov2023segment}
A.~Kirillov, E.~Mintun, N.~Ravi, H.~Mao, C.~Rolland, L.~Gustafson, T.~Xiao,
  S.~Whitehead, A.~C. Berg, W.-Y. Lo, \emph{et~al.}, ``Segment anything,'' in
  \emph{Proceedings of the IEEE/CVF International Conference on Computer
  Vision}, 2023, pp. 4015--4026.

\bibitem{kanazawa2018end}
A.~Kanazawa, M.~J. Black, D.~W. Jacobs, and J.~Malik, ``End-to-end recovery of
  human shape and pose,'' in \emph{Proceedings of the IEEE conference on
  computer vision and pattern recognition}, 2018, pp. 7122--7131.

\bibitem{corral20253darticcyclists}
E.~R. Corral-Soto, Y.~Liu, Y.~Ren, B.~Dongfeng, T.~Cao, and B.~Liu,
  ``3darticcyclists: Generating synthetic articulated 8d pose-controllable
  cyclist data for computer vision applications,'' in \emph{2025 IEEE
  Intelligent Vehicles Symposium (IV)}.\hskip 1em plus 0.5em minus 0.4em\relax
  IEEE, 2025, pp. 2114--2121.

\bibitem{sun2020scalability}
P.~Sun, H.~Kretzschmar, X.~Dotiwalla, A.~Chouard, V.~Patnaik, P.~Tsui, J.~Guo,
  Y.~Zhou, Y.~Chai, B.~Caine, \emph{et~al.}, ``Scalability in perception for
  autonomous driving: Waymo open dataset,'' in \emph{Proceedings of the
  IEEE/CVF conference on computer vision and pattern recognition}, 2020, pp.
  2446--2454.

\bibitem{lin2014microsoft}
T.-Y. Lin, M.~Maire, S.~Belongie, J.~Hays, P.~Perona, D.~Ramanan,
  P.~Doll{\'a}r, and C.~L. Zitnick, ``Microsoft coco: Common objects in
  context,'' in \emph{Computer Vision--ECCV 2014: 13th European Conference,
  Zurich, Switzerland, September 6-12, 2014, Proceedings, Part V 13}.\hskip 1em
  plus 0.5em minus 0.4em\relax Springer, 2014, pp. 740--755.

\bibitem{zuffi2018lions}
S.~Zuffi, A.~Kanazawa, and M.~J. Black, ``Lions and tigers and bears: Capturing
  non-rigid, 3d, articulated shape from images,'' in \emph{Proceedings of the
  IEEE conference on Computer Vision and Pattern Recognition}, 2018, pp.
  3955--3963.

\bibitem{corral2023domain}
E.~R. Corral-Soto, M.~Rochan, Y.~Y. He, X.~Chen, S.~Aich, and L.~Bingbing,
  ``Domain adaptation in lidar semantic segmentation via hybrid learning with
  alternating skip connections,'' in \emph{2023 IEEE Intelligent Vehicles
  Symposium (IV)}.\hskip 1em plus 0.5em minus 0.4em\relax IEEE, 2023, pp. 1--7.

\bibitem{zhang2013rider}
Y.~Zhang, K.~Chen, and J.~Yi, ``Rider trunk and bicycle pose estimation with
  fusion of force/inertial sensors,'' \emph{IEEE Transactions on Biomedical
  engineering}, vol.~60, no.~9, pp. 2541--2551, 2013.

\bibitem{yang2024lemon}
Y.~Yang, W.~Zhai, H.~Luo, Y.~Cao, and Z.-J. Zha, ``Lemon: Learning 3d
  human-object interaction relation from 2d images,'' in \emph{Proceedings of
  the IEEE/CVF Conference on Computer Vision and Pattern Recognition}, 2024,
  pp. 16\,284--16\,295.

\bibitem{hodavn2016evaluation}
T.~Hoda{\v{n}}, J.~Matas, and {\v{S}}.~Obdr{\v{z}}{\'a}lek, ``On evaluation of
  6d object pose estimation,'' in \emph{Computer Vision--ECCV 2016 Workshops:
  Amsterdam, The Netherlands, October 8-10 and 15-16, 2016, Proceedings, Part
  III 14}.\hskip 1em plus 0.5em minus 0.4em\relax Springer, 2016, pp. 606--619.

\bibitem{ahmadyan2021objectron}
A.~Ahmadyan, L.~Zhang, A.~Ablavatski, J.~Wei, and M.~Grundmann, ``Objectron: A
  large scale dataset of object-centric videos in the wild with pose
  annotations,'' in \emph{Proceedings of the IEEE/CVF conference on computer
  vision and pattern recognition}, 2021, pp. 7822--7831.

\end{thebibliography}

\end{document}